# Automatic kidney segmentation in ultrasound images using subsequent boundary distance regression and pixelwise classification networks

Shi Yin*[&], Qinmu Peng*, Hongming Li[&], Zhengqiang Zhang*, Xinge You*, Katherine Fischer[#,♫], Susan L. Furth[§], Gregory E. Tasian[#,♫♪], Yong Fan[&]

*School of Electronic Information and Communications, Huazhong University of Science and Technology, Wuhan, China
[&]Department of Radiology, Perelman School of Medicine, University of Pennsylvania, Philadelphia, PA, 19104, USA
[§]Department of Pediatrics, Division of Pediatric Nephrology, The Children's Hospital of Philadelphia, Philadelphia, PA 19104, USA
[#]Department of Surgery, Division of Pediatric Urology, The Children's Hospital of Philadelphia, Philadelphia, PA 19104, USA
[♫]Center for Pediatric Clinical Effectiveness, The Children's Hospital of Philadelphia, Philadelphia, PA 19104, USA
[♪]Department of Biostatistics, Epidemiology, and Informatics, The University of Pennsylvania, Philadelphia, PA, 19104, USA

**Abstract**: It remains challenging to automatically segment kidneys in clinical ultrasound (US) images due to the kidneys' varied shapes and image intensity distributions, although semi-automatic methods have achieved promising performance. In this study, we propose subsequent boundary distance regression and pixel classification networks to segment the kidneys, informed by the fact that the kidney boundaries have relatively homogenous texture patterns across images. Particularly, we first use deep neural networks pre-trained for classification of natural images to extract high-level image features from US images, then these features are used as input to learn kidney boundary distance maps using a boundary distance regression network, and finally the predicted boundary distance maps are classified as kidney pixels or non-kidney pixels using a pixel classification network in an end-to-end learning fashion. We also adopted a data-augmentation method based on kidney shape registration to generate enriched training data from a small number of US images with manually segmented kidney labels. Experimental results have demonstrated that our method could effectively improve the performance of automatic kidney segmentation, significantly better than deep learning-based pixel classification networks.

# 1. INTRODUCTION

Ultrasound (US) imaging has been widely used to aid diagnosis and prognosis of acute and chronic kidney diseases (Ozmen et al., 2010; Pulido et al., 2014). In particular, anatomic characteristics derived from US imaging, such as renal elasticity, are associated with kidney function (Meola et al., 2016) and lower renal parenchymal area as measured on US imaging data is associated with increased risk of end-stage renal disease (ESRD) in boys with posterior urethral valves (Pulido et al., 2014). Imaging features computed from US data using deep convolutional neural networks (CNNs) improved the classification of children with congenital abnormalities of the kidney and urinary tract (CAKUT) and controls (Zheng et al., 2019; Zheng et al., 2018a). The computation of these anatomic measures typically involves manual or semi-automatic segmentation of kidneys in US images, which increases inter-operator variability and reduces reliability. Therefore, automatic and reliable segmentation of the kidney from US imaging data is desired.

Since manual segmentation of the kidney is time consuming, labor-intensive, and highly prone to intra- and inter-operator variability, semi-automatic and interactive segmentation methods have been developed (Torres et al., 2018). Particularly, an interactive tool has been developed for detecting and segmenting the kidney in 3D US images (Ardon et al., 2015). A semi-automatic segmentation framework based on both texture and shape priors has been proposed for segmenting the kidney from noisy US images (Jun et al., 2005). A graph-cuts method has been proposed to segment the kidney in US images by integrating image intensity information and texture information (Zheng et al., 2018b). A variety of methods have been proposed to segment the kidney based on active shape models and statistical shape models (Ardon et al., 2015; Cerrolaza et al., 2016; Cerrolaza et al., 2014; Martín-Fernández and Alberola-López, 2005; Mendoza et al., 2013). Random forests have also been adopted in a semi-automatic segmentation method to segment the kidney (Sharma et al., 2015). Although a variety of strategies have been adopted in the semi-automatic kidney segmentation methods, most of them solve the kidney segmentation problem as a boundary detection problem and rely on manual operations for initializing the semi-automatic segmentation.

Deep CNNs have demonstrated excellent performance in a variety of image segmentation problems, including semantic segmentation of natural images (Badrinarayanan et al., 2017; Chen et al., 2018a; Chen et al., 2018b; Long et al., 2015; Zhao et al., 2017) and medical image segmentation (Li et al., 2019; Ronneberger et al., 2015; Zhao et al., 2016; Zhao et al., 2018a, b). In these studies, the image segmentation problems are solved as pixelwise or voxelwise pattern classification problems. Recently, several methods have been proposed to automatically segment the kidney from medical imaging data to generate kidney masks using deep CNNs. In particular, Unet networks have been adopted to segment the kidney (Jackson et al., 2018; Ravishankar et al., 2017; Sharma et al., 2017). In these pattern classification-based kidney segmentation methods, all pixels/voxels within the kidney have the same kidney classification labels. Such a strategy might be sensitive to large variability of the kidneys in both appearance and shape in US images. As shown in Fig. 1, kidneys may have varied shapes and heterogeneous appearances in US images. The shape and appearance variability of kidneys, in conjunction with inherent speckle noise of US images, may degrade performance of the pixelwise pattern classification based kidney segmentation methods (Noble and Boukerroui, 2006).

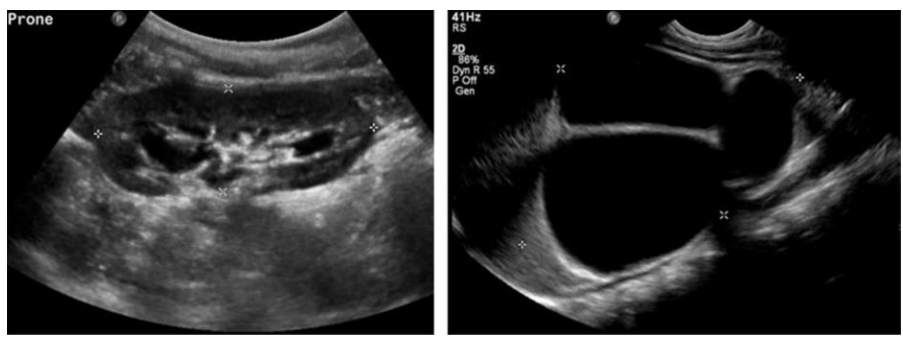

Fig. 1. Kidneys in US images may have varied shapes and the kidney pixels typically have heterogeneous intensities and textures.

On the other hand, several recent studies have demonstrated that pixelwise pattern classification based image segmentation methods could achieve improved image segmentation performance by incorporating boundary detection as an auxiliary task in both natural image segmentation (Bertasius et al., 2015; Chen et al., 2016) and medical image segmentation (Chen et al., 2017; Tang et al., 2018). However, the performance of these methods is hinged on their image pixelwise classification component as the boundary detection serves as an auxiliary task for refining edges of the pixelwise segmentation results.

Inspired by the excellent performance of the semi-automatic boundary detection-based kidney segmentation methods, we develop a fully automatic, end-to-end deep learning method to subsequently learn kidney boundaries and pixelwise kidney masks from a set of manually labeled US images. Instead of directly distinguishing kidney pixels from non-kidney ones in a pattern classification setting, we learn CNNs in a regression setting to detect kidney boundaries that are modeled as boundary distance maps. From the learned boundary distance maps, we learn pixelwise kidney masks by optimizing their overlap with the manual kidney segmentation labels. To augment the training dataset, we adopt a kidney shape-based image registration method to generate more training samples. Our deep CNNs are built upon an image segmentation network architecture derived from DeepLab (Chen et al., 2018b) so that existing image classification/segmentation models could be reused as a starting point of the kidney image segmentation in a transfer learning framework to speed up the model training and improve the performance of the kidney image segmentation. We have evaluated the proposed method for segmenting the kidney based on clinical US images, including 185 images for methodology development and 104 images for performance evaluation. Experimental results have demonstrated that the proposed method could achieve promising segmentation performance and outperformed alternative state-of-the-art deep learning based image segmentation methods, including FCNN (Long et al., 2015), Deeplab (Chen et al., 2018b), SegNet (Badrinarayanan et al., 2017), Unet (Ronneberger et al., 2015), PSPnet (Zhao et al., 2017), and DeeplabV3+(Chen et al., 2018a). Preliminary results of this study will be presented at ISBI 2019 (Yin et al., 2019).

## 2. METHODS AND MATERIALS

### 2.1 Imaging Data

All the imaging data used in this study were collected using standard clinical US scanners at the Children's Hospital of Philadelphia (CHOP). Particularly, 185 first post-natal kidney US images in sagittal view were obtained from our previous studies (Zheng et al., 2019; Zheng et al., 2018b), including 85 from 50 children with congenital abnormalities of the kidney and urinary tract (one from each abnormal kidney), and 100 from 50 children with unilateral mild hydronephrosis (one from each kidney). We randomly selected 105 images as training data and 20 images as validation data in the present study to train deep learning-based kidney segmentation models. The kidney segmentation models were then evaluated using the remaining 60 images and another set of 104 first post-natal US images in sagittal view that were also obtained from the CHOP. All the images were obtained from different kidneys. The images were manually segmented by experts from the CHOP. Representative kidney US images are shown in Fig.1.

These images were resized to have the same size of 321×321, and their image intensities were linearly scaled to [0,255].

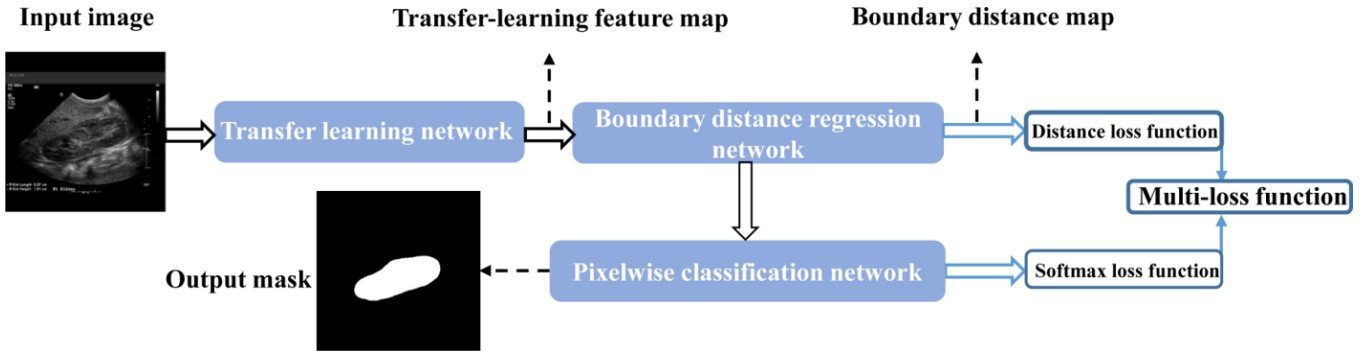

Fig. 2. Transfer learning network, and subsequent boundary distance regression and pixel classification networks for fully automatic kidney segmentation in US images.

### 2.2 Deep CNN Networks for Kidney Image Segmentation

The kidney image segmentation method is built upon deep CNNs to subsequently detect kidney boundaries and kidney masks in an end-to-end learning fashion. As illustrated in Fig.2, the kidney image segmentation model consists of a transfer learning network, a boundary distance regression network, and a kidney pixelwise classification network. The transfer learning network is built upon a general image classification network to reuse an image classification model as a starting point for learning high level image features from US images, the boundary distance regression network learns kidney boundaries modeled as distance maps of the kidney boundaries, and the kidney boundary distance maps are finally used as input to the kidney pixelwise classification network to generate kidney segmentation masks.

The kidney distance regression network and the kidney pixelwise classification network are trained based on augmented training data that are generated using a kidney shape-based image registration method. The network architecture and the data augmentation methods are described in following sections.

*2.2.1 Transfer learning network for extracting high level image features from US images*

Instead of directly building an image segmentation network on raw US images, we adopt a transfer learning strategy to extract informative image features from US images as a starting point for learning high level image features from US images. Particularly, we extract features from US images by utilizing a general image classification network, namely VGG16, which achieves 92.7% top-5 test-accuracy in ImageNet [34].

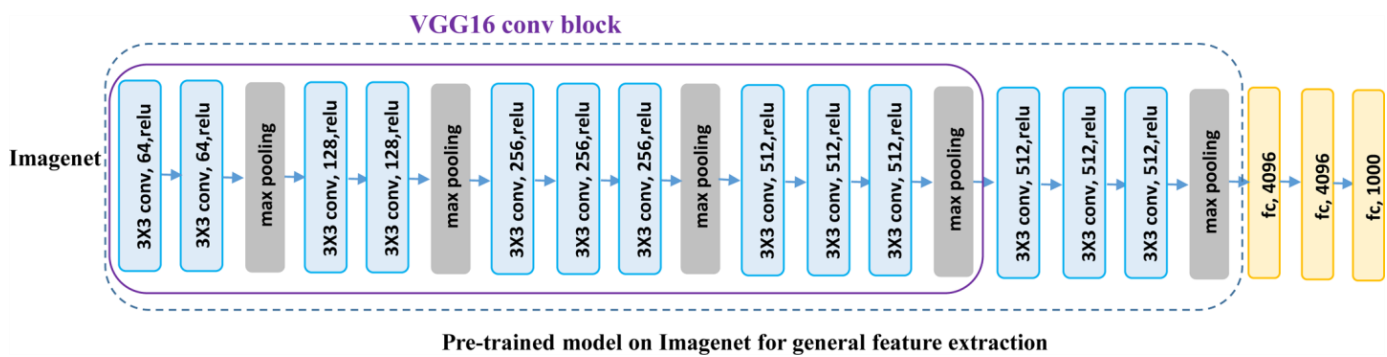

Fig. 3. The architecture of theVGG16 model.

As illustrated in Fig. 3, the VGG16 network consists of 16 convolutional (conv) layers with a receptive field of 3×3. The stack of convolutional layers is followed by 3 fully-connected (FC) layers: each of the first two has 4096 channels and the third performs 1000-way classification. The final layer is a softmax layer. In our experiments, we fine-tuned the model weights of Imagenet-pretrained VGG16 network to adapt them to the boundary distance regression network.

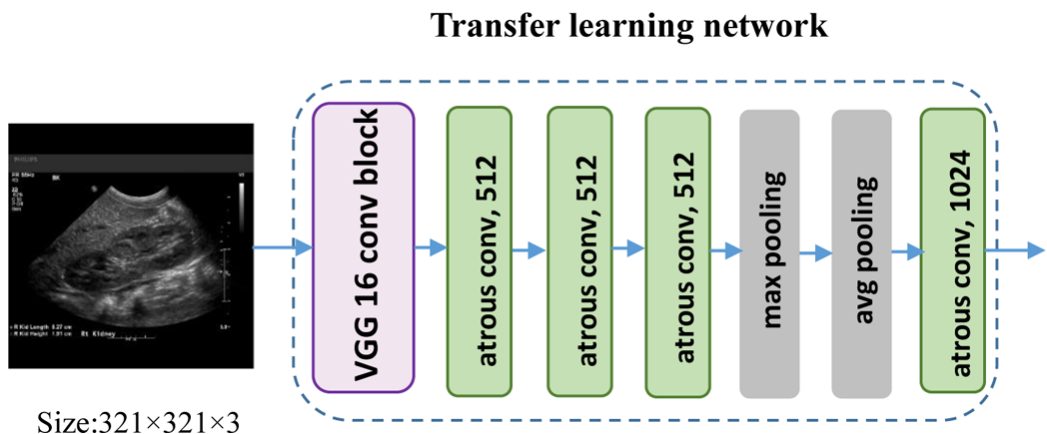

Fig. 4. The architecture of the Deeplab model. We extracted the pretrained feature maps L from the exiting Deeplab model.

On the other hand, we follow the Deeplab architecture (Chen et al., 2018b) by applying the atrous convolutions to compute denser image feature representations. As illustrated in Fig. 4, the DeepLab image segmentation model modified the last 3 convolutional layers and FC layers into 4 atrous convolutional layers and 2 convolutional classification layers. We discarded the 2 convolutional classification layers to adapt the architecture to the boundary distance regression network.

*2.2.2 Boundary distance regression network for fully automatic kidney segmentation in ultrasound images*

We develop a boundary distance regression network for fully automatic kidney segmentation in ultrasound images, instead of directly learning a pixelwise classification network directly from the US image features because the heterogenous kidney appearances in US images make it difficult to directly classify pixels as kidney or background pixels.

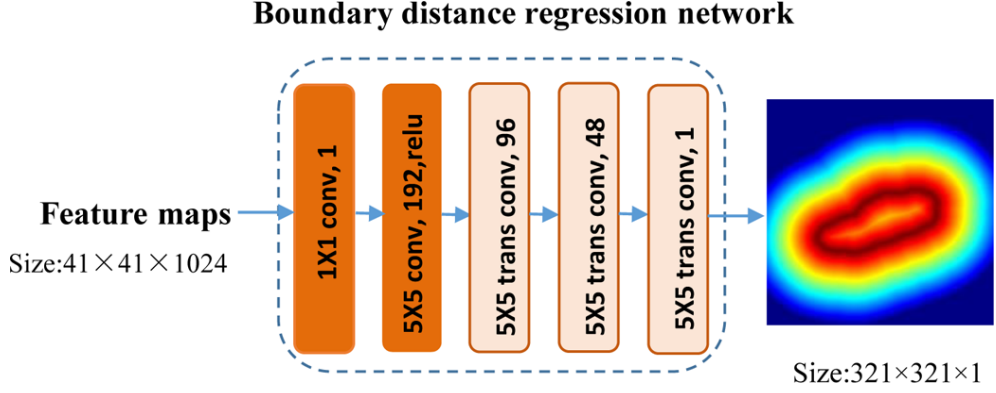

Fig. 5. Network architecture of the boundary distance regression network.

The boundary distance regression network is built in a regression framework, consisting of two parts: a projection part that produces boundary feature maps $S_0$ and a high-resolution reconstruction part that upsamples the feature maps to obtain the kidney distance maps at the same spatial resolution of the input images as illustrated in Fig. 5. The projection part is built on two convolutional layers, and the reconstruction part is built on deconvolution (transposed convolution) layers. The output of the *i*th deconvolution operation $S_i$ is defined as

$$S_i = max(0, W_i \otimes S_{i-1} + B_i), \tag{1}$$

where $W_i$ is deconvolution filters with size $f_i \times f_i$, and $B_i$ is a bias vector, and $\otimes$ is deconvolution operator. The upsampling deconvolution layers double the spatial dimension of their input feature maps, and therefore 3 upsampling deconvolution layers are adopted in the kidney boundary distance regression network to learn the kidney boundary in the input image space. In our experiments, the numbers of input feature maps of the upsampling deconvolution layers were empirically set to 3 times of 64, 32, 16 respectively, and the filter size $f_i$ was empirically set to 5.

We solve the kidney boundary detection problem in a regression framework to learn distance from the kidney boundary at every pixel of the input US images. The kidney boundary detection problem could be potentially solved in an end-to-end classification framework (Xie and Tu, 2015). However, the number of the kidney boundary pixels is much smaller than the number of non-boundary pixels in US kidney images. Such unbalanced boundary and non-boundary pixels make it difficult to learn an accuracy classification model. Therefore, we model the kidney boundary detection as a distance map learning problem.

Given an input US image $I$ with its kidney boundary, we compute the distance to the kidney boundary for every pixel $P_i \in I$ of the input image and obtain a normalized kidney distance map of the same size of the input image using potential function as following:

$$d(P_i) = exp(-\lambda D_i),  \qquad (2)$$

with $D_i = \min_{b_j \in \boldsymbol{b}} \text{dist}(P_i, b_j)$ is the minimal Euclidean distance of pixel $P_i$ to the kidney boundary pixels $\boldsymbol{b} = \{b_j\}_{j \in J}$, and $\lambda$ is a parameter. As illustrated in Fig.6, at the kidney boundary pixels the normalized exponential kidney distance equals to 1. In this study, the kidney boundary is detected by learning the normalized kidney distance map.

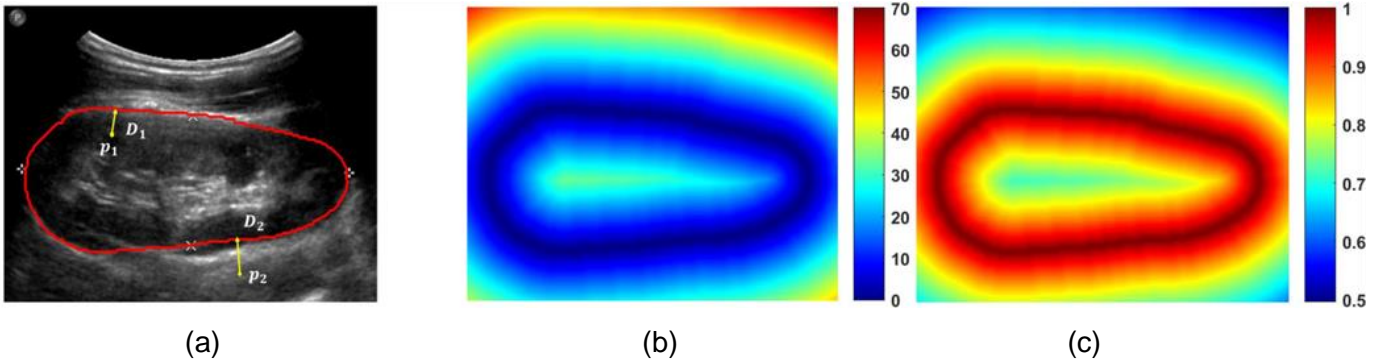

(a) (b) (c)

Fig. 6. An example kidney US image and kidney boundary (a), its boundary distance map (b), and its normalized potential distance map with $\lambda = 1$ (c). The colorbar of (c) is in log scale.

To learn the normalized kidney boundary distance map, we train the boundary detection network by minimizing a distance loss function $L_d$, defined as,

$$L_d = \sum_{P_i \in I} \|\varphi(P_i) - d(P_i)\|^2, \qquad (3)$$

where $\varphi(P_i)$ is the predicted distance and $d(P_i)$ is the ground truth distance to the kidney boundary at pixel $P_i$. Once we obtain the predicted distance for every pixel of a US image to be segmented, we can obtain a boundary binary map with a threshold $e^{-\lambda}$.

To obtain a smooth closed contour of the kidney boundary, we construct a minimum spanning tree of all predicted kidney boundary pixels. Particularly, we first construct an undirected graph of all predicted kidney boundary pixels, each pair of which are connected with a weight of their Euclidean distance. Then, a minimum spanning tree *T* is obtained using Kruskal's algorithm (Kruskal, 1956). Finally, the max path of the minimum spanning tree *T* is obtained as a close contour of the kidney boundary and a binary mask of the kidney is subsequently obtained. To reduce the complexity of finding a minimum spanning tree, we could apply morphological operations to the binary kidney boundary image to obtain a skeleton binary map and apply the minimum spanning tree algorithm to the skeleton binary map. We refer to *the boundary distance regression network followed by post-processing for segmenting kidneys* as a boundary detection network hereafter.

*2.2.3 Subsequent boundary distance regression and pixelwise classification networks for fully automatic kidney segmentation in ultrasound images*

The minimum spanning tree post-processing method could obtain promising boundary detection results for most US kidney images. However, it fails if the predicted boundary distance maps are far from perfect. To obtain robust kidney segmentation performance, we propose to learn pixelwise kidney masks from the predicted kidney distance maps.

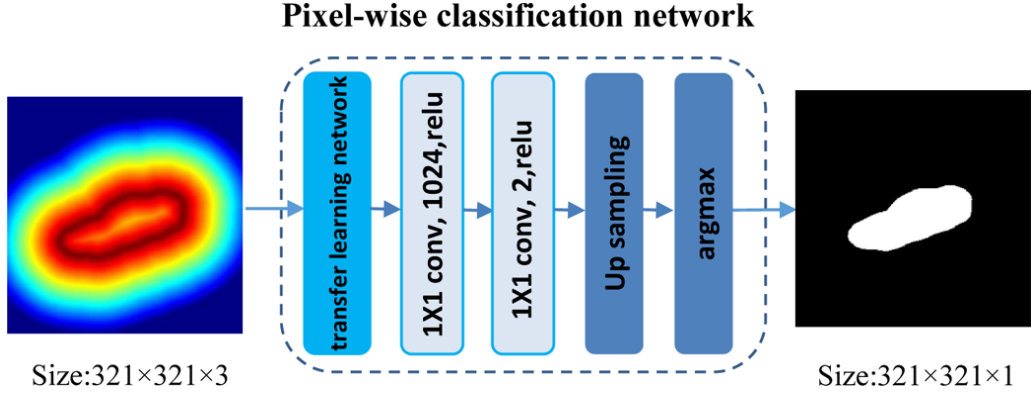

Fig. 7. Architecture of the kidney pixel classification network.

As illustrated in Fig. 7, the kidney pixelwise classification network is built upon a semantic image segmentation network, namely Deeplab image segmentation network. The input to the kidney pixel classification network is an image with 3 channels, each being the output of the predicted kidney boundary map. The final decision for class labels is then made by applying 2-channel classification softmax layer to the extracted class feature maps $f_{(c,P_i)}$ ($c$=0,1) based on cross-entropy loss function. We also adopt the pre-trained VGG16 model to initialize the pixelwise classification network parameters.

To train the kidney boundary distance regression and pixelwise classification networks in an end-to-end fashion, loss functions of the kidney boundary distance regression network and the kidney pixelwise classification network are combined with a multi-loss function

$$L_m = \left(1 - \frac{\tau}{N}\right) L_d + \frac{\tau}{N} \gamma L_s, \qquad (5)$$

where $N$ is the number of total iterative training steps, $\tau$ is an index of the training iteration step, and $\gamma$ is a parameter to make $L_d$ and $L_s$ to have the same magnitude and to be determined empirically. The multi-loss function puts more weight on the kidney boundary distance regression cost function in the early stage of the network training and then shifts to the kidney pixel classification cost function in the late stage of the network training. The kidney pixelwise classification network's output is treated as the overall segmentation result.

## 2.3 Data augmentation based on kidney shape registration

To build a robust kidney segmentation model, we augment the training data using a non-rigid image registration method (Bookstein, 1989). Particularly, in order to generate training US kidney images with varied kidney shapes and appearances, we register each training image to all other training images based on thin-plate splines transformation (TPS) (Bookstein, 1989) as following.

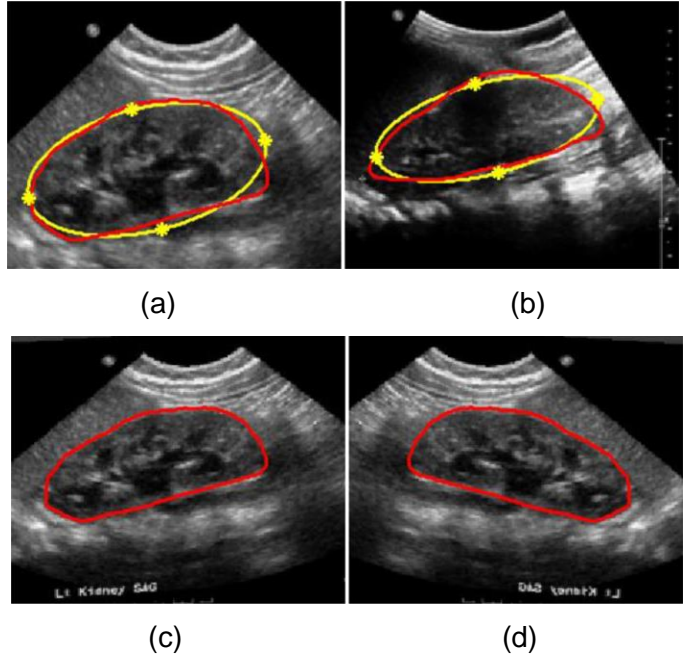

(a)                   (b)

(c)                   (d)

Fig. 8. Data-augmentation based on TPS transformation and flipping. (a) is a moving image, (b) is a fixed image, (c) is the registered image and (d) is the flipped registered image. The kidney shape denoted by the red curve is approximately modeled as an ellipse denoted by the yellow curves. The yellow stars denote the landmark points of the TPS transformation.

Given two US kidney images with kidney boundaries, we register one image (moving image $M$) to the other image (fixed image $F$), in order to generate a deformed moving image in the fixed image space. First, the kidney boundary is approximately modeled as an ellipse as shown in Fig. 8. Then, ellipse vertexes of the kidney are identified as 4 corresponding landmark points across kidneys, and the landmark points of the fixed and moving images are denoted by $Z_M = [x_M\ y_M]^T \in R^{2\times 4}$ and $Z_F = [x_F\ y_F]^T \in R^{2\times 4}$ respectively. Base on the corresponding landmark points, a TPS operation **W** is computed to register the moving image $M$ to the fixed image $F$, defined as:

$$W = \begin{bmatrix} R & P^T \\ P & 0^{3\times 3} \end{bmatrix}^{-1} \begin{bmatrix} Z_M^T \\ 0^{3\times 2} \end{bmatrix}, \quad (6)$$

where $P = [I^{4\times 1}\ Z_M^T]^T \in R^{3\times 4}$ is the homogeneous coordinates of $Z_M$, and R is a symmetric matrix with elements $r_{i,j} = \phi(\|Z_M - Z_M\|_2)$. Based on the estimated TPS transformation, a warped moving image could be generated. We use a nearest neighbor interpolation method to warp the moving image. In addition to the image registration-based data augmentation, we also flip the training images left to right. Given $n$ training images, we can obtain $2n(n-1) + n$ augmented training images with kidney boundaries.

## 2.4 Implementation and Evaluation of the proposed method

The proposed networks were implemented based on Python 3.7.0 and TensorFlow r1.11. We used a mini-batch of 20 images to train the deep neural networks. The maximum number of iteration steps was set to 20000. The deep learning models were trained on a GeForce GTX 6.0GB graphics processing unit (GPU). The deep networks were optimized using Adam stochastic optimization (Boyd and Vandenberghe, 2004) with the learning rate of $10^{-4}$. Besides the parameters of transfer learning network, the filters of the boundary regression network were initialized with random normal initializer with the mean 0 and standard deviation 0.001. The biases of the boundary regression network were initialized with constant 0. In the subsequent kidney boundary detection and pixel classification network, γ was empirically set to 1 to make the kidney boundary regression and pixel classification loss functions to have similar magnitudes. The minimum spanning tree algorithm were implemented based on Networkx python library (https://networkx.github.io/documentation/networkx-1.10/overview.html).

Ablation studies were carried out to evaluate how different components of the proposed method affect the segmentation performance. We first trained the kidney boundary detection network using following 3 different strategies, including training from scratch without data augmentation (named "random+noaug"), transfer learning without data augmentation (named "finetune+noaug"), and transfer-learning with data augmentation (named "finetune+aug"). Particularly, for the training from scratch we adopted "Xavier" initialization method that has been widely adopted in natural image classification and segmentation studies (Glorot and Bengio, 2010). Particularly, we initialized the biases to be 0 and filters to follow a uniform distribution. The training images' normalized kidney boundary distance maps were obtained with $\lambda = 1$. Outputs of the kidney boundary detection network were post-processed using morphological operations and minimum spanning tree algorithm in order to obtain kidney masks.

Second, we trained and compared kidney boundary detection networks based on augmented training data and transfer-learning initialization with their normalized kidney distance maps obtained with different values of λ, including 0.01, 0.1, 1, and 10. We also trained a kidney boundary detection network in a classification setting. Particularly, the classification-based kidney boundary detection network had the same network architecture as the regression based kidney boundary detection network, except that its loss function was a softmax cross-entropy loss to directly predict the kidney boundary pixels. Outputs of the kidney boundary detection network were also post-processed using the morphological operations and minimum spanning tree algorithm in order to obtain kidney masks.

Third, we trained and compared kidney boundary detection networks with 3 state-of-the-art pixel classification based image segmentation deep neural networks, namely FCNN (Long et al., 2015), Deeplab (Chen et al., 2018b), SegNet (Badrinarayanan et al., 2017), Unet (Ronneberger et al., 2015), PSPnet (Zhao et al., 2017), and DeeplabV3+(Chen et al., 2018a). These networks also adopted encoding part of the VGG16, the same one as adopted in our method. The comparisons were based on both the training data without augmentation and the augmented training data. Segmentation performance of all the methods under comparison was measured using Dice index, mean distance (MD), Jaccard, Precision, Sensitivity, ASSD defined in (Hao et al., 2014; Zheng et al., 2018c). The segmentation performance measures of testing images obtained by different segmentation networks were statistically compared using Wilcoxon signed-rank tests (Woolson, 2008).

We also compared the proposed subsequent boundary distance regression and pixelwise classification networks with the kidney boundary detection network. We trained the subsequent boundary distance regression and pixelwise classification networks based on the augmented training data and transfer-learning initialization. Finally, we compared the proposed framework with a multi-task learning based segmentation network to jointly estimate the kidney distance maps and classify the kidney pixels. We obtained the results in 20000 iteration numbers. Particularly, the multi-task learning based segmentation network adopted the same network architecture of the kidney boundary distance regression network and included a branch for the kidney pixel classification.

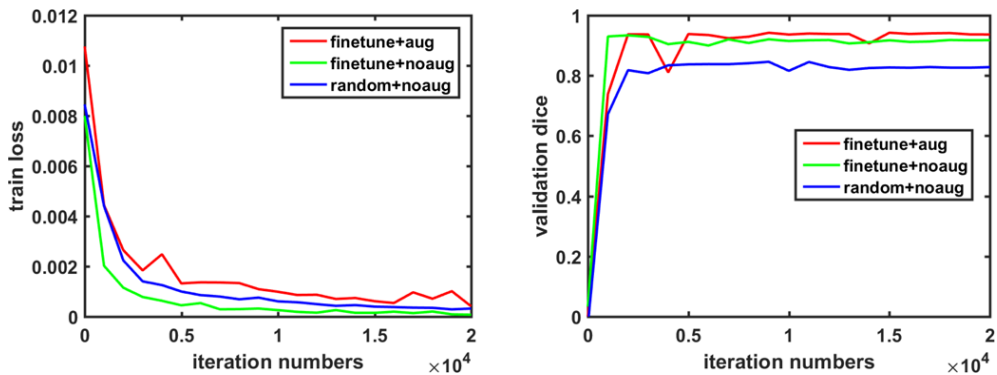

Fig. 9. Traces of the training loss (left) and validation accuracy (right) associated with 3 different training strategies.

## 3. EXPERIMENTAL RESULTS

Sections 3.1, 3.2, and 3.3 summarize segmentation performance of the kidney boundary detection network (the boundary distance regression network followed by post-processing) and its comparison with pixelwise classification segmentation methods, including FCNN (Long et al., 2015), Deeplab (Chen et al., 2018b), SegNet (Badrinarayanan et al., 2017), Unet (Ronneberger et al., 2015), PSPnet (Zhao et al., 2017), and DeeplabV3+(Chen et al., 2018a). Section 3.4 summarizes segmentation performance of the subsequent boundary distance regression and pixelwise classification networks and these networks integrated under a multi-task framework.

**3.1 Effectiveness of transfer-learning and data-augmentation on the segmentation performance**

Fig. 9 shows traces of the training loss and validation accuracy of the kidney boundary detection network trained with 3 different training strategies. These traces demonstrate that the training of the kidney boundary distance regression network converged regardless of the training strategies used. Without the data augmentation, the transfer-learning strategy made the model to better fit the training data and obtained better segmentation accuracy than the random initialization strategy. Although the train loss was relatively larger if the network was trained based on the augmented data, better validation segmentation accuracy was obtained. Validation segmentation accuracy measures of different training strategies are summarized in Table I and example segmentation results are illustrated in Fig. 10, further demonstrating that the transfer learning and data augmentation strategies could make the boundary detection network to achieve better results. The results were obtained at 9000, 1000, 16000 iteration numbers respectively with the different training strategies according to their validation accuracy traces.

Table 1. Segmentation performance of different training strategies on the validation dataset.

| Methods | random+noaug | finetune+noaug | finetune+aug |
|---|---|---|---|
| Dice | 0.8458±0.1714 | 0.9338±0.0377 | 0.9421±0.0343 |
| MD | 6.1173±4.7427 | 3.2896±2.0465 | 3.0804±2.7781 |

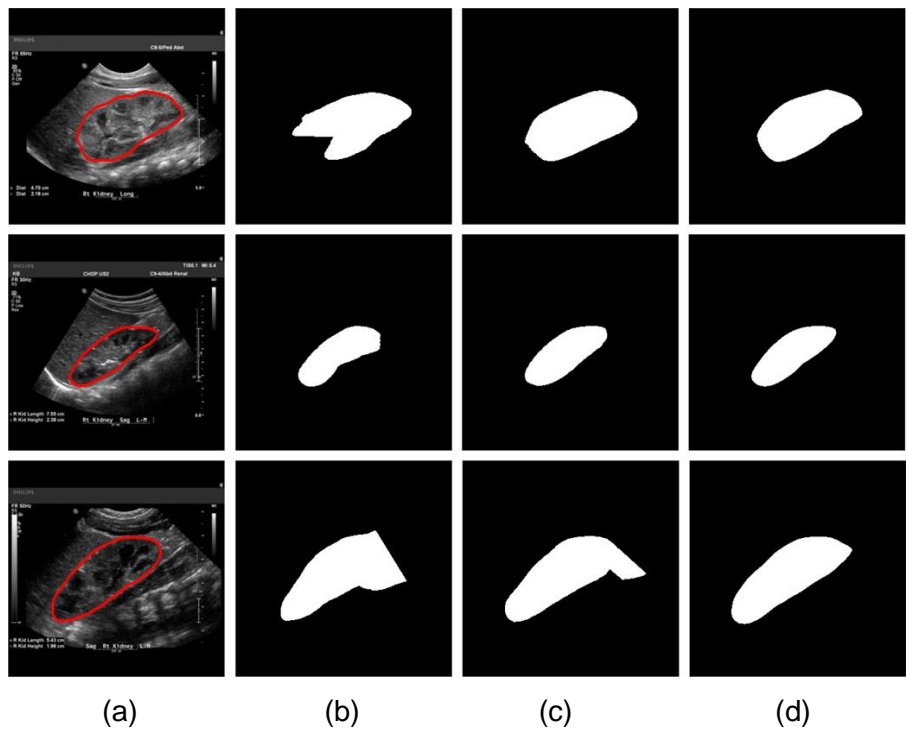

(a)　　　　(b)　　　　(c)　　　　(d)

Fig. 10. Example segmentation results obtained with 3 training strategies. (a) input image and ground truth boundary, (b) results of the training from scratch without data augmentation, (c) results of the transfer learning without data augmentation, and (d) results of the transfer-learning with data augmentation.

Table 2. Segmentation results of the boundary detection network trained with distance functions with different settings on the validation dataset.

|  | $\lambda = 0.01$ | $\lambda = 0.1$ | $\lambda = 1$ | $\lambda = 10$ |
|---|---|---|---|---|
| Dice | 0.9003±0.0978 | 0.9364±0.0501 | 0.9421±0.0343 | 0.9264±0.0593 |
| MD | 4.2726±3.4268 | 2.8683±1.9569 | 3.0804±2.7781 | 3.3180±2.7279 |

**3.2 Performance of the kidney boundary detection networks trained using different loss functions**

Fig.11 shows example kidney boundaries and masks obtained by the boundary detection networks trained with different distance maps in the validation dataset. Particularly, the kidney masks were closer to the ground truth when the distance maps were normalized with $\lambda = 1$ than other values. When $\lambda = 0.01$, the detected boundaries were much border than the ground truth, while when $\lambda = 10$ or $\lambda = 0.1$, the detected boundaries missed some pixels. Such a problem could be overcome to some extent by the post-processing steps, including morphological operations and minimum spanning tree. As summarized in Table II, the kidney segmentation results obtained with λ = 0.1 and λ = 1 were comparable, better than those obtained with other parameters. It is worth noting that predicted boundary distance regression results shown in Fig. 11 were not binary maps when λ = 1 or 10. We set λ = 1 in all following experiments. We also found that the softmax cross-entropy loss function based pixelwise classification network classified all the pixels into background due to the unbalanced samples.

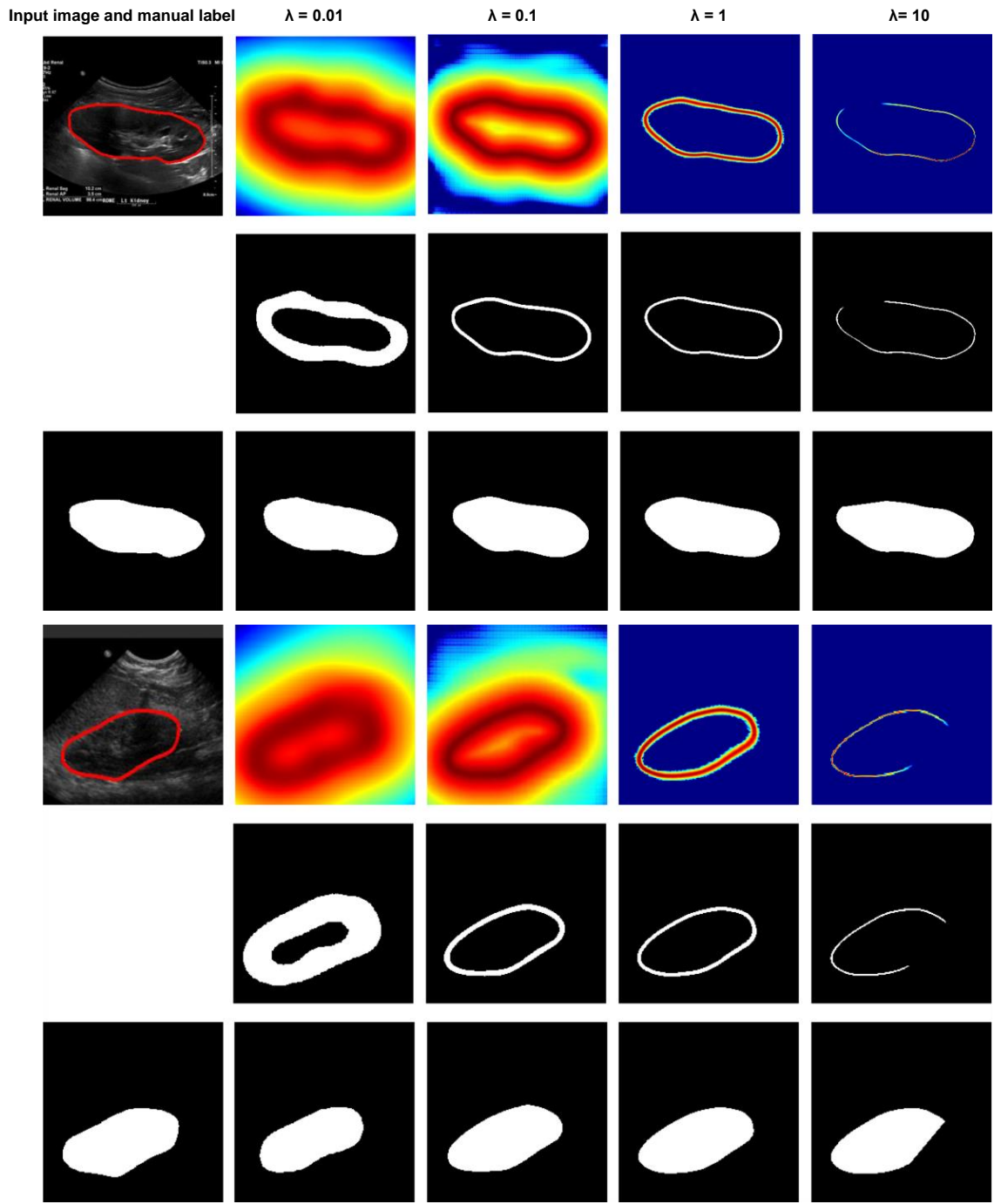

Fig. 11. Results of the kidney boundary detection networks trained using different loss functions. The 1st and 4rd rows show predicted boundary distance maps, the 2st and 5rd rows show boundary binary maps, and the 3nd and 6th rows show kidney masks obtained by the morphology operation and minimum spanning tree based post-processing method.

Table 3. Comparison results for the proposed method and other methods without data augmentation on the testing dataset.

|  | Dice coefficient | | Mean distance | | Jaccard | | Sensitivity | | Precision | | ASSD | |
| --- | --- | --- | --- | --- | --- | --- | --- | --- | --- | --- | --- | --- |
|  | Mean±std | p-value | Mean± std | p-value | Mean± std | p-value | Mean±std | p-value | Mean± std | p-value | Mean±std | p-value |
| FCNN | 0.8020±0.1037 | 3.1e-27 | 9.6083±4.8746 | 4.1e-26 | 0.6806±0.1316 | 1.4e-27 | 0.7614±0.1404 | 1.1e-25 | 0.8690±0.1118 | 1.7e-15 | 9.9992±5.4128 | 1.2e-27 |
| Deeplab | 0.8724±0.0639 | 8.4e-26 | 6.2596±2.9087 | 2.0e-21 | 0.7788±0.0899 | 7.1e-26 | 0.8759±0.1002 | 3.2e-12 | 0.8829±0.0788 | 2.9e-19 | 6.3022±3.1650 | 8.8e-25 |
| SegNet | 0.8894±0.0619 | 2.9e-20 | 7.3921±6.1747 | 2.2e-23 | 0.8061±0.0949 | 1.6e-20 | 0.8809±0.0917 | 5.4e-9 | 0.9080±0.0814 | 2.3e-10 | 6.1979±4.6609 | 4.8e-21 |
| Unet | 0.8899±0.0578 | 1.7e-20 | 7.5227±4.1757 | 8.8e-26 | 0.8062±0.0883 | 1.2e-20 | 0.8780±0.0900 | 4.7e-12 | 0.9121±0.0745 | 5.8e-09 | 6.2147±3.3764 | 7.9e-24 |
| PSPnet | 0.8869±0.0548 | 2.0e-22 | 7.2514±4.0461 | 3.2e-23 | 0.8008±0.0828 | 1.0e-22 | 0.8818±0.0885 | 6.1e-10 | 0.9027±0.0741 | 4.4e-16 | 6.0621±3.0109 | 4.7e-22 |
| Deeplabv3+ | 0.8965±0.0698 | 2.4e-13 | 6.4669±3.8065 | 1.9e-18 | 0.8187±0.1014 | 2.0e-13 | 0.8643±0.1100 | 9.6e-13 | 0.9424±0.0474 | - | 5.3100±2.9109 | 5.7e-16 |
| Proposed | 0.9303±0.0499 | - | 3.6130±2.3447 | - | 0.8729±0.0708 | - | 0.9220±0.0690 | - | 0.9442±0.0482 | - | 3.4792±1.9618 | - |

Table 4. Comparison results for the proposed method and other methods with data augmentation on the testing dataset.

|  | Dice coefficient | | Mean distance | | Jaccard | | Sensitivity | | Precision | | ASSD | |
| --- | --- | --- | --- | --- | --- | --- | --- | --- | --- | --- | --- | --- |
|  | Mean±std | p-value | Mean± std | p-value | Mean± std | p-value | Mean±std | p-value | Mean± Std | p-value | Mean±std | p-value |
| FCNN | 0.9179±0.0471 | 4.8e-17 | 4.2072±2.4922 | 4.2e-16 | 0.8514±0.0731 | 2.4e-17 | 0.9221±0.0730 | 4.8e-06 | 0.9207±0.0622 | 6.0e-10 | 4.1182±2.3157 | 1.5e-16 |
| Deeplab | 0.9236±0.0370 | 1.9e-18 | 3.7369±1.8737 | 4.5e-14 | 0.8601±0.0603 | 1.3e-18 | 0.9175±0.0584 | 3.0e-15 | 0.9344±0.0528 | 1.9e-03 | 3.7779±1.8596 | 1.6e-16 |
| SegNet | 0.9298±0.0409 | 2.3e-11 | 4.0753±6.4316 | 3.6e-10 | 0.8713±0.0651 | 1.6e-11 | 0.9311±0.0542 | 2.1e-05 | 0.9330±0.0600 | 2.0e-04 | 3.8450±4.9187 | 8.9e-10 |
| Unet | 0.9245±0.0403 | 2.7e-14 | 4.0987±2.5463 | 5.8e-14 | 0.8620±0.0650 | 2.6e-14 | 0.9132±0.0621 | 3.3e-13 | 0.9423±0.0614 | 4.3e-02 | 3.8643±2.1498 | 1.1e-14 |
| PSPnet | 0.9372±0.0305 | 1.8e-05 | 3.3055±2.0240 | 2.7e-04 | 0.8837±0.0580 | 3.4e-04 | 0.9491±0.0440 | - | 0.9292±0.0577 | 1.0e-09 | 3.1471±1.6704 | 1.7e-03 |
| DeeplabV3+ | 0.9311±0.0354 | 2.8e-08 | 3.5080±2.2860 | 5.0e-05 | 0.8732±0.0610 | 1.7e-05 | 0.9307±0.0517 | 1.6e-05 | 0.9369±0.0612 | 2.1e-03 | 3.1837±1.9855 | 8.1e-05 |
| Proposed | 0.9427±0.0453 | - | 2.9189±1.9823 | - | 0.8943±0.0663 | | 0.9439±0.0645 | | 0.9458±0.0413 | - | 2.8701±1.8671 | |

### 3.3 Segmentation performance of kidney boundary detection and pixel classification networks

Table 3 and Table 4 show kidney segmentation results of the testing data obtained by FCNN, Deeplab, SegNet, Unet, PSPnet, deeplabV3+, and the boundary detection network (Bnet), trained without the data augmentation or with the data augmentation. The morphology operation and minimum spanning tree based post-processing method was used to obtain kidney masks from the boundary detection network. The results demonstrate that the boundary detection network had significantly better performance than the alternative deep learning segmentation networks that were trained to classify pixels into kidney and non-kidney pixels. The results also demonstrated that the data-augmentation could improve the performance of all the methods under comparison. Fig. 12 shows representative segmentation results obtained by the deep learning methods under comparison with or without the data augmentation. These results demonstrate that our method had robust performance regardless of the variance of kidney shape and appearance. However, the alternative methods under comparison had relatively worse performance.

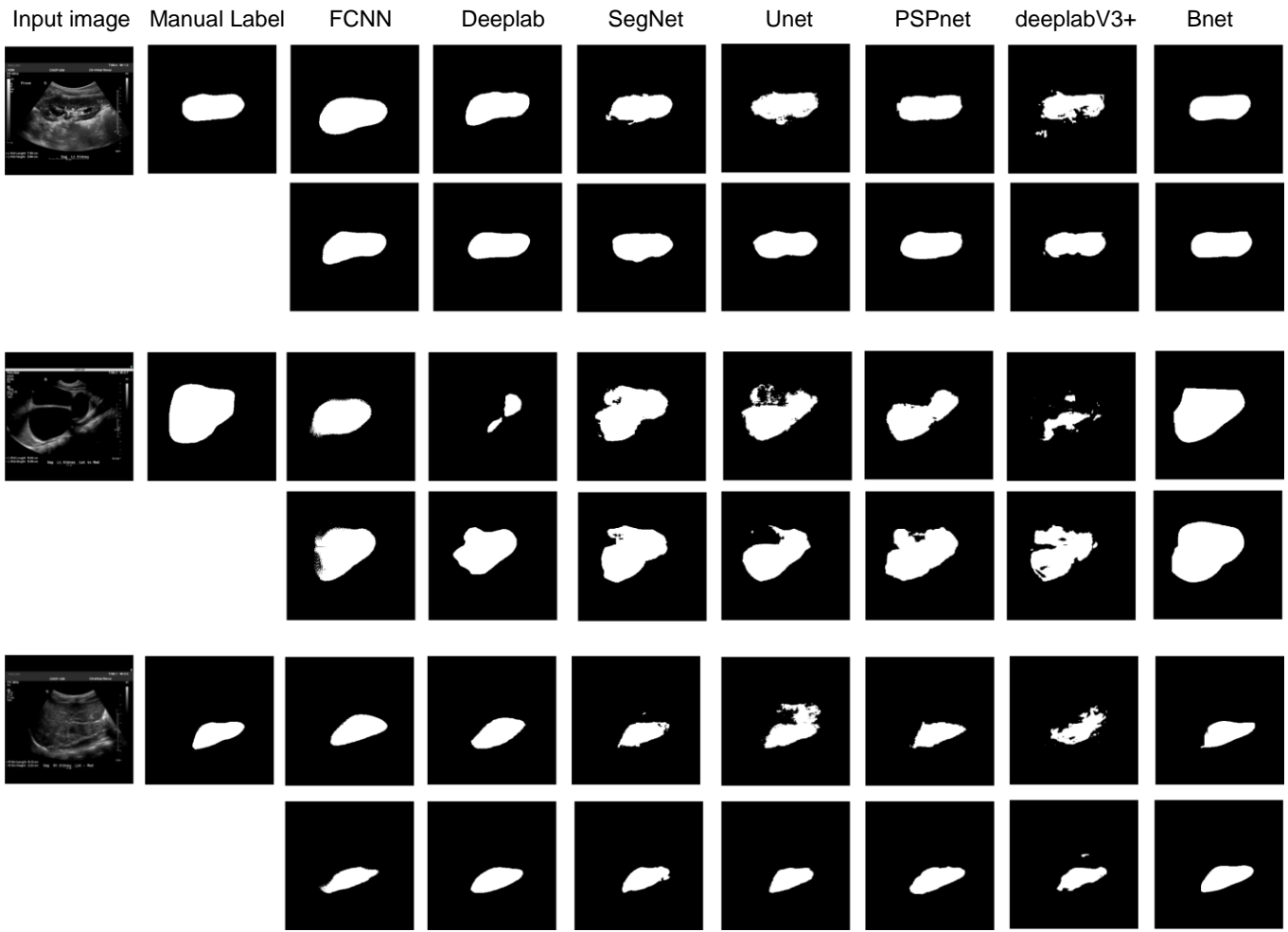

Fig. 12. Representative segmentation results obtained by different deep learning networks trained without data augmentation (the 1$^{st}$, 3$^{nd}$, and 5$^{rd}$ rows) and with data augmentation (the 2$^{th}$, 4$^{th}$, and 6$^{th}$ rows).

Table 5. Comparison results for the proposed model with only distance loss function and end-to-end subsequent segmentation framework on the testing dataset.

|  | Boundary detection network | End-to-end learning |
| --- | --- | --- |
| Dice | 0.9427±0.0453 | 0.9451±0.0315 |
| MD | 2.9189±1.9823 | 2.6822±1.4634 |
| Time (sec) | 3.75 | 0.18 |

### 3.4 Comparison of the end-to-end subsequent segmentation network, the boundary detection network, and the multi-task learning based segmentation network

Table 5 shows segmentation performance of the kidney boundary detection network and the end-to-end subsequent boundary distance regression and pixelwise classification networks, demonstrating that the end-to-end learning could obtain better performance, although the difference was not statistically significant. Fig. 13 indicated that the proposed framework could improve the boundary detection segmentation results with blurring boundaries. More importantly, we can get the kidney masks from their distance maps without any post-processing step. As illustrated by intermediate results shown in Fig. 13 (d), the end-to-end network obtained distance maps as expected. A further comparison of computational time costs of two solutions indicated that the end-to-end learning was 20 times faster than the morphology and minimum spanning tree based post-processing.

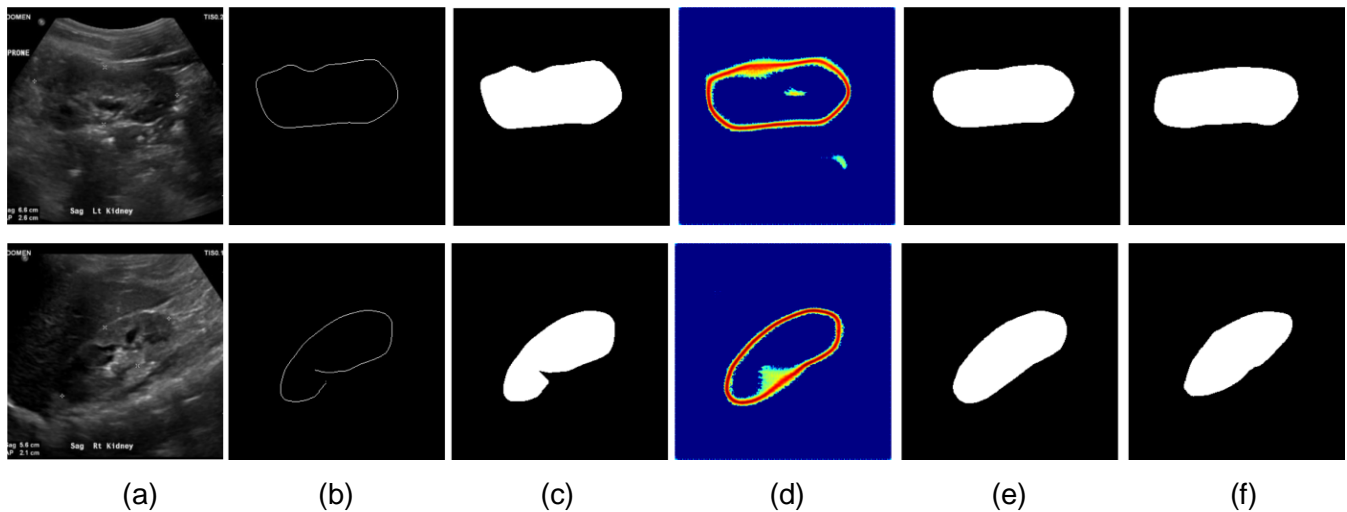

  (a)  (b)  (c)  (d)  (e)  (f)

Fig. 13. Results for the boundary detection network and the end-to-end learning networks. (a) input kidney US images, (b) binary skeleton maps of the predicted distance maps, (c) kidney masks obtained with the minimum spanning tree based post-processing, (d) predicted distance maps obtained by the end-to-end subsequent segmentation network, (e) kidney masks obtained by the end-to-end subsequent segmentation network, and (f) kidney masks obtained by manual labels.

  We also combined the proposed boundary distance regression network and the Deeplab mask segmentation network in a multi-task learning framework. In the multi-task learning framework, we set the weight parameter of the two networks to 1. The dice coefficient of segmentation results obtained by the boundary detection network was 91.19%±5.69% and the dice coefficient of segmentation results obtained by the mask segmentation network was 93.14%±3.45%. We found that accuracy of the mask segmentation network was improved in the multi-task learning framework. However, we also found that the accuracy of the boundary detection network was reduced, partially due to the inaccurate results of the mask segmentation networks.

## 4. DISCUSSION

In this study, we propose a novel boundary distance regression network architecture to achieve fully-automatic kidney segmentation, and the boundary distance regression network is integrated with a subsequent pixelwise classification network to achieve improved kidney segmentation performance in an end-to-end learning fashion, instead of directly learning a pixelwise classification neural network to distinguish kidney pixels from non-kidney ones. Our method has been applied to segmentation of clinical kidney US images with large variability in both appearance and shape, yielding promising performance, similar to a semi-automatic method (Zheng et al., 2018b) and significantly better than the alternatives under comparison. Our results have demonstrated that the boundary detection strategy worked better than pixelwise classification techniques for segmenting clinical US images that have demonstrated excellent image segmentation performance in a variety of image segmentation problems, including FCNN (Long et al., 2015), Deeplab (Chen et al., 2018b), SegNet (Badrinarayanan et al., 2017), Unet (Ronneberger et al., 2015), PSPnet (Zhao et al., 2017), and DeeplabV3+(Chen et al., 2018a). Our results have also demonstrated that the kidney shape registration-based data-augmentation method could improve the segmentation performance.

We found that the network initialization was important for the pixelwise classification based deep learning kidney image segmentation methods. In all our experiments, all the pixel-classification based image segmentation models were trained in a transfer learning framework by adopting the encoding part of the VGG16, the same one as adopted in our method. Without the transfer learning part, these pixel-classification based deep learning kidney image segmentation models often failed to converge in our experiment, probably due to that the clinical kidney images had heterogeneous appearances in our study, as illustrated by Fig. 1. Such shape and appearance variability, in conjunction with inherent speckle noise of US images, may degrade performance of the pixelwise pattern classification-based kidney segmentation methods. In contrast, the boundary distance regression network successfully converged with the same random initialization. These results might indicate that for the segmentation of clinical kidney US images with varied shapes and appearances the boundary distance regression is a better strategy than the pixel classification strategy.

Instead of adopting random transformations to generate more US kidney images, we used an image registration method to augment the training image data so that the augmented imaging data have meaningful realistic kidney shapes. The results shown in Table 3 and Table 4 have demonstrated that the data-augmentation method could improve all the deep learning-based segmentation methods under comparison in terms of the segmentation accuracy.

The benefits of the end-to-end learning for the kidney segmentation are twofold. First, the end-to-end learning could improve the computational efficiency due to the usage of GPUs. Second, the segmentation performance could be further improved, as indicated by the results shown in Table 5 and Fig. 13. We found that the improvement was significant for kidney images with blurry boundaries, although the average performance difference was not statistically significant.

The segmentation results of the multi-task learning network with integrated boundary distance regression network and Deeplab based pixelwise classification network indicated that the boundary distance regression network could improve the performance of the pixelwise classification network. However, a better strategy might be needed to further improve its performance.

Though our kidney segmentation network is built on the pre-trained VGG16 network, our model is not limited to the VGG network. Within the same framework, we could also adopt other semantic segmentation networks with good performance in natural images segmentation tasks, such as ResNet (He et al., 2016) or network built on VGG19 (Simonyan and Zisserman, 2014), which may further improve the performance of the kidney segmentation results. Moreover, our method could be further improved by incorporating multi-scale learning strategies that have demonstrated helpful in image segmentation (Chen et al., 2018a; Zhao et al., 2017).

# 5. CONCLUSIONS

We have developed a fully automatic method to segment kidneys in clinical ultrasound images by integrating boundary distance regression and pixel classification networks subsequently in an end-to-end learning framework. Experimental results have demonstrated that the boundary distance regression network could robustly detect boundaries of kidneys with varied shapes and heterogeneous appearances in clinical ultrasound images, and the end-to-end learning of subsequent boundary distance regression and pixel classification networks could effectively improve the performance of automatic kidney segmentation, significantly better than deep learning-based pixel classification networks.

# ACKNOWLEDGEMENTS

This work was supported in part by the National Institutes of Health (DK117297 and DK114786); the National Center for Advancing Translational Sciences of the National Institutes of Health (UL1TR001878); the National Natural Science Foundation of China (61772220 and 61473296); the Key Program for International S&T Cooperation Projects of China (2016YFE0121200); the Hubei Province Technological Innovation Major Project (2017AAA017 and 2018ACA135); the Institute for Translational Medicine and Therapeutics' (ITMAT) Transdisciplinary Program in Translational Medicine and Therapeutics, and the China Scholarship Council.